\documentclass{article}
\usepackage{arxiv}
\usepackage[utf8]{inputenc}
\usepackage[T1]{fontenc}
\usepackage{hyperref}
\usepackage{url}
\usepackage{cite}
\usepackage{amsmath,amssymb,amsfonts}
\usepackage{algorithmic}
\usepackage{graphicx}
\usepackage{textcomp}
\usepackage{multirow}
\usepackage{float}

\def\BibTeX{{\rm B\kern-.05em{\sc i\kern-.025em b}\kern-.08em
    T\kern-.1667em\lower.7ex\hbox{E}\kern-.125emX}}

\title{X-GuideAR: An Augmented Reality Framework to Mitigate Radiation Exposure during Fluoroscopic Guidance. A use case in S2AI Screw Placement.}

\author{
  Mingxu Liu\thanks{Corresponding author: mliu90@jh.edu. This work is partially supported by NIH grants R01EB023939 and R01AR080315.} \\
  Department of Computer Science\\
  Johns Hopkins University\\
  Baltimore, MD, 21211, USA \\
  \And
  Zixuan Liu \\
  Department of Computer Science\\
  Johns Hopkins University\\
  Baltimore, MD, USA \\
  \And
  Ruchen Cai \\
  Department of Computer Science\\
  Johns Hopkins University\\
  Baltimore, MD, USA \\
  \And
  Yu-Chun Ku \\
  Department of Computer Science\\
  Johns Hopkins University\\
  Baltimore, MD, USA \\
  \And
  Suxi Gu \\
  Department of Orthopaedic Surgery\\
  Johns Hopkins University\\
  Baltimore, MD, USA \\
  \And
  Amit Jain \\
  Department of Orthopaedic Surgery\\
  Johns Hopkins School of Medicine\\
  Baltimore, MD, USA \\
  \And
  Alejandro Martin-Gomez\thanks{Joint senior author. Member, IEEE. Department of Electrical Engineering and Computer Science, University of Arkansas, Fayetteville, AR, USA, and Department of Computer Science, Johns Hopkins University, Baltimore, MD, USA.} \\
  Department of Electrical Engineering and Computer Science\\
  University of Arkansas\\
  Fayetteville, AR, USA \\
  \And
  Mehran Armand\thanks{Joint senior author. Senior Member, IEEE. Department of Mechanical Engineering, University of Arkansas, Fayetteville, AR, USA, and Departments of Orthopaedic Surgery and Mechanical Engineering, Johns Hopkins University, Baltimore, MD, USA.} \\
  Department of Mechanical Engineering\\
  University of Arkansas\\
  Fayetteville, AR, USA \\
}

\hypersetup{
pdftitle={X-GuideAR: An Augmented Reality Framework to Mitigate Radiation Exposure during Fluoroscopic Guidance},
pdfauthor={Mingxu Liu, Zixuan Liu, Ruchen Cai, Yu-Chun Ku, Suxi Gu, Amit Jain, Alejandro Martin-Gomez, Mehran Armand},
pdfkeywords={Medical system, virtual reality and interfaces, computer vision for medical robotics},
}

\begin{document}
\maketitle

\begin{abstract}
Achieving optimal screw placement for orthopedic surgeries requires frequent alignment checks and multiple anatomical views under X-ray—a process known as “fluoro-hunting” that increases radiation exposure to patients and surgical teams. This work introduces \textit{X-GuideAR}, an augmented reality (AR) framework for identifying optimal X-ray views, aimed at reducing radiation exposure while ensuring accurate screw placement. To exemplify the benefits of X-GuideAR, we focus on S2 alar-iliac (S2AI) screw placement. Our system provides radiation-free guidance for view acquisition and drilling by generating synthetic X-ray previews that accelerate fluoro-hunting. Once the required anatomical views are identified using these previews, a real X-ray is acquired, and the preview of the drilling trajectory is augmented onto it, facilitating precise screw placement with minimal additional radiation. A preliminary study involving eight S2AI trajectories performed by an expert spine surgeon demonstrated a 62.3\% reduction in the number of X-rays. Post-procedure evaluations showed that trajectories done with \textit{X-GuideAR} supported an average safe screw diameter of 12.95 mm compared to 5.9 mm under the conventional workflow, suggesting improved bony containment and potential biomechanical benefit. \textit{X-GuideAR} shows great potential to reduce radiation exposure and streamline S2AI screw placement, offering a promising direction toward safer and more efficient surgeries.
\end{abstract}


\section{Introduction}
\label{sec:introduction}
Since 1970s, X-rays have been instrumental in orthopedic surgeries to provide visual reference for screw placement with minimizing the invasion, enhancing placement accuracy and reducing complications \cite{wilson2024image}. X-ray devices are among the most proficient and accessible tools for intra-operative guidance of orthopedic surgeries \cite{otomo2022computed,wilson2024image}. Despite its clinical proficiency and accessibility, X-rays are two-dimensional images, superimpose anatomical structures, which can obscure tool positioning relative to critical anatomy and necessitate scans from multiple angles for clarity \cite{wilson2024image}. To obtain the necessary detail, surgeons engage in a process known as fluoro-hunting, adjusting the X-ray device to acquire specific anatomical views that reveal landmarks essential to the procedure \cite{kochanski2019image}. Using these targeted X-ray views, surgeons align their instruments to avoid damaging vital structures while guiding them along the intended trajectory. Achieving precise alignment requires frequent C-arm adjustments and numerous X-ray exposures. Moreover, the number of X-rays needed and the overall surgical outcome heavily depend on the surgeon’s experience with the procedure. Frequent use of X-rays also raises significant concerns about ionizing radiation exposure, posing long-term risks to patients and healthcare providers \cite{kaplan2016intraoperative}.

\begin{figure}[h]
\centering
\includegraphics[width=\columnwidth]{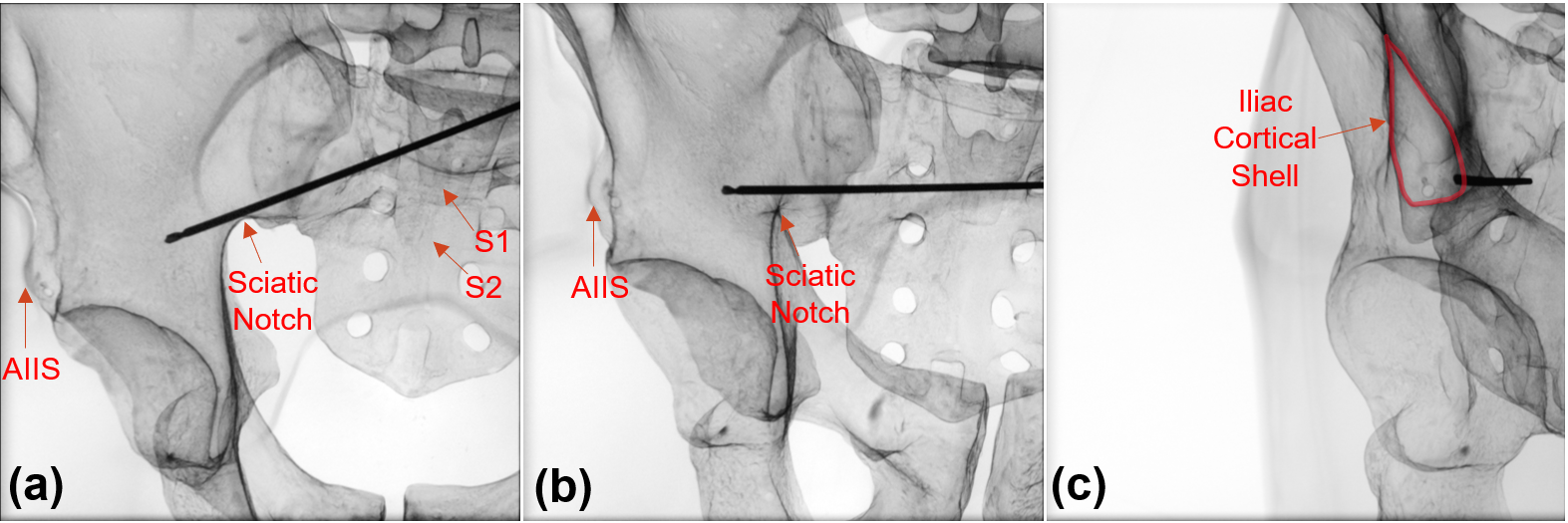}
\caption{Key views for the S2AI procedure: In the anterior-posterior (AP) view, the surgeon identifies an entry point between the S1 and S2 joints, avoiding the sciatic notch and aiming toward the anterior inferior iliac spine (AIIS) (a). The inlet view clarifies the pathway is above the sciatic notch and towards AIIS (b). The teardrop view ensures the pathway ends inside the cortical shell (c).}\label{fig:apinletteardrop}
\end{figure}

These concerns are especially critical in complex procedures like S2 alar-iliac (S2AI) screw placement. This procedure represents an advancement in sacropelvic fixation, addressing the biomechanical limitations of conventional iliac screws and reducing revision rates \cite{chang2009low,ilyas2015comparison}. S2AI screw placement demands surgeons to rely on multiple non-orthogonal views to guide the tool from the S1-S2 joint toward the anterior inferior iliac spine (see Fig. \ref{fig:apinletteardrop}) \cite{orchowski2006use}. This trajectory traverses a complex anatomical region where the surgeon must avoid neurovascular structures, cortical bone violations, and misplacement of the screw \cite{arora2022challenges}. 

Achieving this level of care requires frequent verification of alignment using X-rays, exposing the patient and surgical team to repeated radiation throughout the procedure.

To mitigate radiation exposure and enhance precision and safety in orthopedic surgeries, computer-assisted methods such as volumetric guidance, imageless navigation, and robotic surgery have been explored \cite{sugano2003computer}. Specifically for S2AI screw placement, these methods include free-hand techniques, patient-specific templates, intra-operative computed tomography (iCT), and robotic assistance. Free-hand techniques are radiation-free and widely accessible but are highly dependent on surgeon skill \cite{park2015free, shillingford2018free}. Patient-specific templates improve accuracy and reduce radiation exposure but require 3D-printed guides from patient-specific instrument services or comparable manufacturing and design expertise, which prolongs the pre-operative planning process \cite{zhou2020three, hsu2025current, matsukawa2020accuracy}. iCT enhances navigation accuracy by providing real-time 3D imaging but increases patient radiation exposure and adds substantial cost \cite{phan2017novel, sargut2022intraoperative, van2021comparison}. Robotic assistance offers improved precision and reduced radiation; however, its high cost and limited accessibility remain significant barriers to broader adoption \cite{ShillingfordJamalN2018HvRA, laratta2018robotic}. Despite their benefits, the adoption of these techniques is constrained by the need for specialized equipment, extensive training, or reliance on expert surgical skill. Therefore, in clinical practice, fluoroscopy remains the most popular option for the majority of spine surgeons and medical centers \cite{yilmaz2018s2, ShillingfordJamalN2018HvRA, lee2021accuracy}.

Alternatively, advancements in augmented reality (AR) have significantly enhanced surgical guidance in orthopedic procedures \cite{maier2017surgical}. AR improves the surgeon's understanding of anatomy and reduces radiation exposure during spinal and pelvic surgeries \cite{longo2021augmented, avrumova2023augmented}. For example, Foley et al.\ tracked the X-ray device and tool poses, augmenting fluoroscopic images with the predicted insertion trajectory to minimize radiation exposure during tool alignment \cite{foley2001virtual}. Navab et al. introduced a semi-transparent overlay within the C-arm imaging system, providing real-time fusion of X-ray images with live video of the operative field \cite{navab2009camera}. Fischer et al.\ demonstrated that AR reduces radiation exposure during K-wire placement by overlaying live video or 3D RGB-D views during fluoroscopic guidance, offering safer and more efficient alternatives to conventional workflows \cite{fischer2016preclinical}. These studies demonstrate reductions in mental workload for tool alignment and anatomical interpretation, as well as reductions in radiation during the intervention once the anatomy is clearly visualized. However, optimal view selection still depends on the surgeon’s technique, and the fluoro-hunting process continues to be a primary source of radiation exposure \cite{yamada2018safe}.

To reduce radiation during fluoro-hunting, studies have explored the use of Digitally Reconstructed Radiographs (DRRs) to minimize radiation. DRRs are synthetic X-rays generated from 3D computed tomography (CT) data and are commonly employed for surgical planning and registration \cite{1525180}. In this context, De Silva et al. introduced FluoroSim, a system that utilizes DRRs combined with mechanically encoded C-arm movements to expedite fluoro-hunting for standard pelvic views \cite{de2018virtual}. However, conventional DRRs lack the detail of real fluoroscopic images and often too computationally heavy for intra-operative use \cite{unberath2018augmented}. To address these challenges, Unberath et al. proposed DeepDRR, an improved DRR method leveraging deep learning to segment volumes and performs hardware-accelerated, fast simulation of realistic X-ray formation based on material properties \cite{mathias2018deepdrr}. Leveraging DeepDRRs for fluoro-hunting, Killeen et al. successfully reduced the number of X-rays required to obtain anatomical views from 2–4 scans to just one per view \cite{killeen2023mixed}. These approaches require robotic C-arms for execution and do not reduce radiation exposure during the tool alignment or intervention.
 
More recently, advancements in AR systems using head-mounted devices (HMDs) have enhanced spatial awareness in the operating environment, allowing for more complex guidance and expanding the scope of applications. Aaskov et al. evaluated an AR system that overlays spinal X-rays onto a patient's back, enhancing visualization for spinal procedures \cite{aaskov2019x}. Fotouhi et al. presented a frustum paradigm that displayed X-ray and imaging geometry in situ within the 3D surgical scene, easing hand-eye coordination demands and further decreasing radiation exposure \cite{fotouhi2019interactive}. Johnson et al. demonstrated that HMDs improve surgical ergonomics by providing real-time, hands-free visualization of fluoroscopic images directly presented in the surgeon’s field of view \cite{johnson2022visualization}.

Despite these advantage of using HMDs for surgeries, efficient and accurate surgical scene tracking remains a challenge for utilizing HMDs in interventional settings. Early studies achieved tracking relying on optical trackers and their associated fiducial markers and careful calibration, which increases system complexity and are hindered by line-of-sight limitations \cite{chen2015development}. Other approaches like Vuforia or ArUco simplifies workflows but often produces translational errors exceeding 1 cm, rendering them unsuitable for surgeries that demand high accuracy \cite{costa2024assignment}. Recent approaches have explored the benefits of using AR HMDs sensors to enable tracking of retro-reflective markers, with an improvement in flexibility but still face limitations, including line-of-sight constraints, restricted workspaces, and the requirement for additional lighting \cite{gsaxner2021inside,kunz2020infrared}. Recently, Martin-Gomez et al. introduced STTAR, a system that utilizes the HoloLens 2’s infrared camera and self-localization capabilities to achieve sub-millimeter tracking accuracy. STTAR effectively mitigates line-of-sight issues, getting rid of the registration of space between multiple systems, which improves the flexibility of HMD surgical workflows \cite{martin2023sttar}.

For the S2AI procedure, initial applications of AR have introduced the idea of using HMDs for screw insertion without using fluoroscopy. Heining et al. evaluated the early feasibility of using HMDs to visualize and align trajectories for pelvic screw placement, and reported a mean entry point deviation of 3.45 mm for S2AI screws \cite{heining2024augmented}. Judy et al. conducted a human study using AR-HMDs to guide screw placement by displaying CT slices and navigation through the HMD, demonstrating surgical performance comparable to traditional methods \cite{judy2023human}. However, without fluoroscopy, these approaches lacked real-time anatomical feedback, potentially impacting surgical accuracy and undermining surgeon confidence near critical anatomy. 

Alternatively, registration-based navigation systems for screw placement are also widely explored. These systems register intra-operatively reconstructed anatomy—typically acquired using structured light—with pre-operative CT data, enabling image rendering from the CT and facilitating tool alignment \cite{huang2020augmented, kalfas2021machine, lim2023machine}. While registration-based methods have been shown to reduce intra-operative radiation exposure, they require additional surgeon training for proper use, often rely on high-cost specialized equipment, and tend to increase overall surgical duration \cite{virk2019navigation}. Furthermore, intra-operative fluoroscopy remains necessary for placement verification and percutaneous screw insertion \cite{kalfas2021machine, lim2023machine}.

We aim to establish a surgeon-friendly and readily adoptable paradigm that reduces radiation exposure for both the surgical team and the patient throughout the procedure. In this work, we present \textit{X-GuideAR}, an AR-based surgical copilot system designed to assist in orthopedic screw insertion, with a specific focus on S2AI screw placement. \textit{X-GuideAR} facilitates efficient fluoro-hunting to acquire critical views and provides radiation-free tool guidance overlaid on these views. This surgical copilot system leverages an HMD for spatial awareness and DeepDRRs for fast and realistic X-ray preview generation. The key contributions of this work are as follows: (i) \textit{X-GuideAR} minimizes radiation exposure by generating realistic DeepDRR previews, enabling prompt C-arm fluoro-hunting for essential views. (ii) \textit{X-GuideAR} augments fluoroscopic imaging with predicted insertion outcomes to improve trajectory awareness, with the potential to facilitate workflow efficiency through reduced view acquisition imaging. (iii) The safety and feasibility of \textit{X-GuideAR} are validated through experiments conducted in collaboration with an experienced orthopedic surgeon. To the best of our knowledge, this study is the first to employ AR technology to enhance fluoroscopic guidance for the S2AI procedure.
\section{Methodology}\label{sec2}
\begin{figure}[ht]
\centering
\includegraphics[width=\columnwidth]{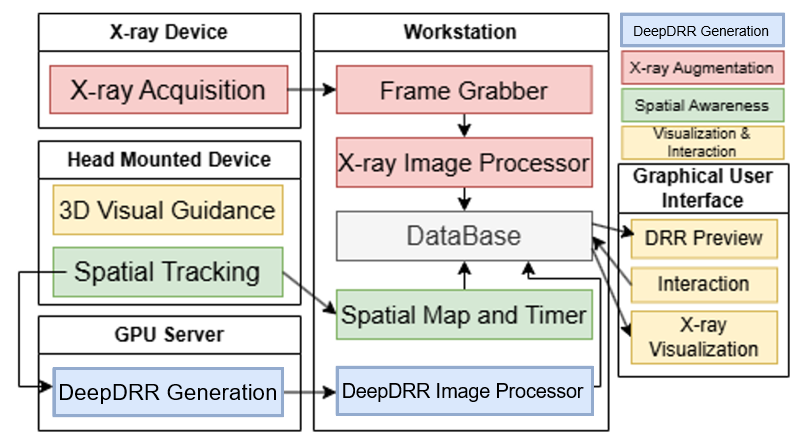}
\caption{\textit{X-GuideAR} integrates four core modules across multiple hardware components: the workstation and X-ray device for X-ray augmentation (Section~\ref{sec:X-ray Augmentation Module}), the HMD for spatial tracking and AR visualization (Sections~\ref{spatialawareness} and~\ref{section:xguidearS2AI}), the GPU server for DeepDRR generation (Section~\ref{section:drr generation}), and the workstation for interaction and system control (Section~\ref{section:xguidearS2AI}).}\label{fig:systemdiagram}
\end{figure}

\noindent \textit{X-GuideAR} comprises four core modules that communicate wirelessly via a publisher-subscriber system (Fig. \ref{fig:systemdiagram}): (I) A \textit{Spatial Awareness} module continuously tracks and maintains the spatial relationship of critical surgical components, such as patient, surgeon, or X-ray device, establishing the kinematic chain essential for the functioning of the remaining three modules. (II) A \textit{DeepDRR Generation} module provides realistic X-ray previews during the fluoro-hunting phase, facilitating precise C-arm positioning. (III) \textit{X-ray Augmentation} module overlays virtual tool trajectories onto X-ray scans, enhancing intra-operative visualization. (IV) A \textit{Visualization and Interaction} module developed in Unity, which includes: (a) a 3D guidance system on the HMD that show the field-of-view of X-ray and the preview trajectory of the surgical tool, extended from the concept introduced in \cite{fotouhi2019interactive}, and (b) a graphical user interface that visualizes DeepDRR previews, displays real-time X-ray augmentations, and enables user interaction with the system.
\subsection{Spatial Awareness Module}\label{spatialawareness}
\noindent Retro-reflective markers are commonly used for navigation in orthopedic surgeries. Recent advancements in inside-out tracking on HMDs offer capabilities comparable to outside-in systems, but with increased flexibility and self-localization \cite{martin2023sttar}. The integration of this tracking algorithm can accurately localize surgical markers attached to the C-arm, patient, and surgical tool, denoted as $T_{S}^{C}$, $T_{S}^{P}$, and $T_{S}^{T}$ (see Fig. \ref{fig:surgicalscene}). To fully establish the kinematic chain necessary for precise navigation, additional steps involving calibration and registration are required. Specifically:

\begin{figure}[ht]
\centering
\includegraphics[width=\columnwidth]{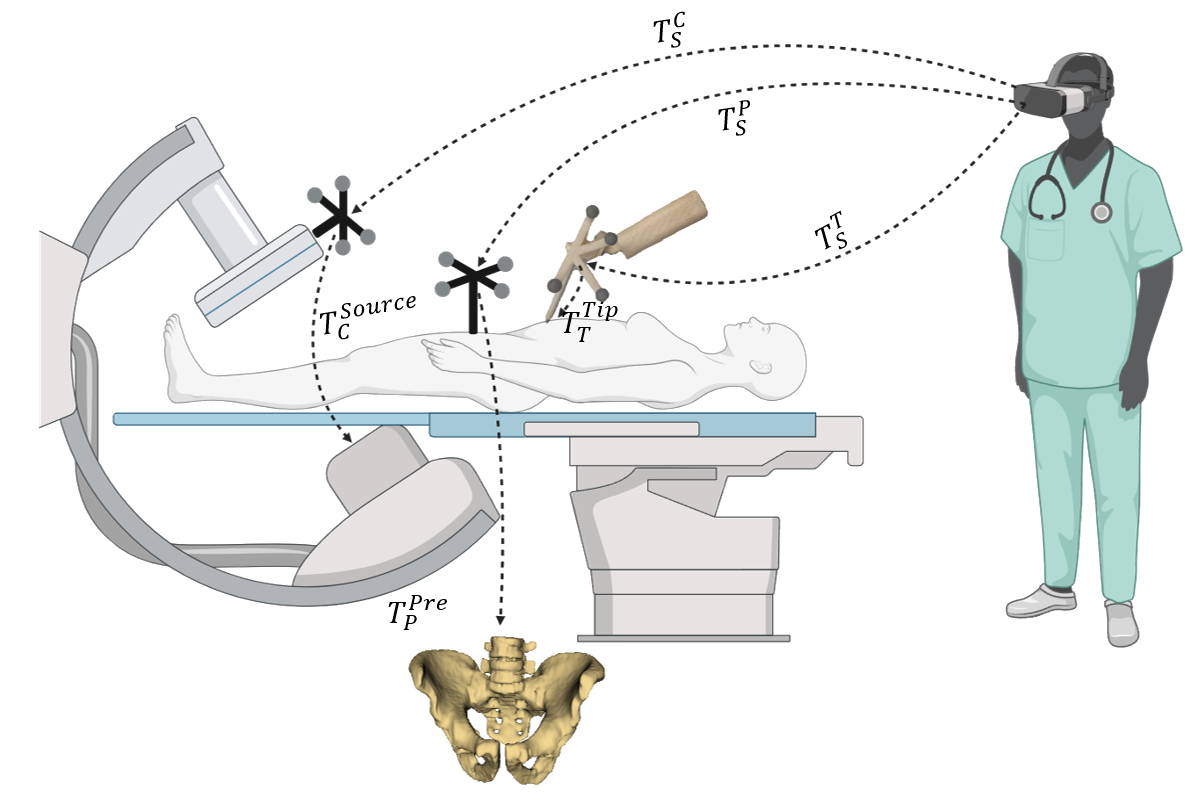}
\caption{The AR system maintains a spatial map containing surgeon, X-ray device, patient anatomy, and surgical tools. (Created with BioRender.com.)}\label{fig:surgicalscene}
\end{figure}

\begin{figure*}[t!]
\centering
\includegraphics[width = 1\textwidth]{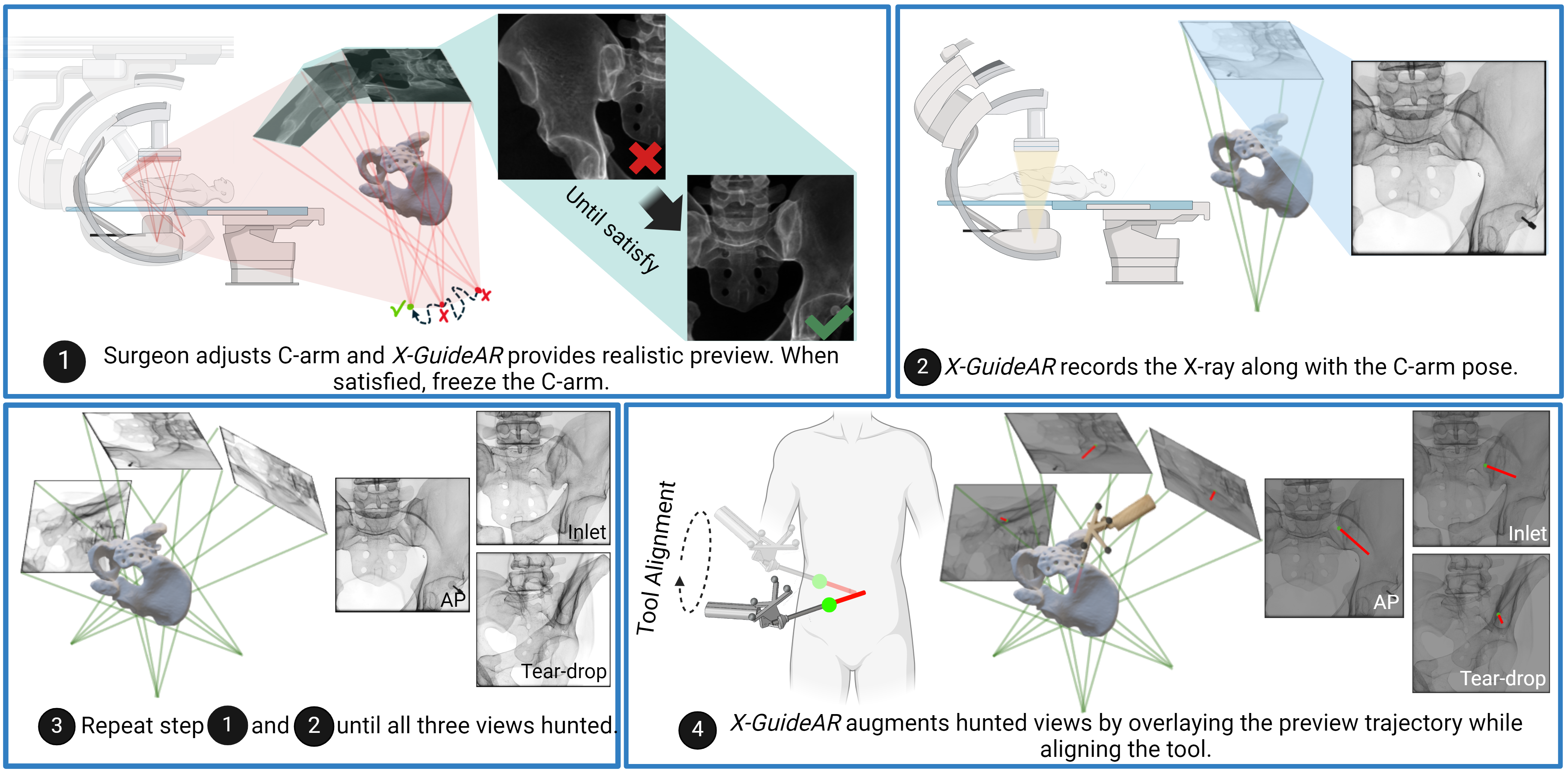}
\caption{\textit{X-GuideAR} aids in the S2AI procedure by providing real-time previews for fluoro-hunting and screw navigation. The fluoro-hunting preview reduces radiation exposure during anatomical view acquisition, while the screw navigation preview minimizes radiation during drill alignment. Created with BioRender.com.}
\label{fig:systemoverview}
\end{figure*}

\subsubsection{X-ray Source}\label{section:X-raysource}
\noindent The X-ray device, typically a C-arm in orthopedic surgeries, can be tracked by rigidly attaching markers to the gantry. Determining the pose of the C-arm source—which is crucial for DeepDRR generation—requires calibration of the C-arm. In this work, we calibrate the C-arm using the method described in \cite{fotouhi2019interactive}, which involves solving a Perspective-n-Point (PnP) problem. This is a one-time step-up, which does not affect the workflow of surgery.

By modeling the C-arm source as a pinhole camera, its intrinsic matrix $K$ can be obtained from the engineering manual if factory calibration is available, or estimated through an additional calibration procedure \cite{hosseinian2019toward}. A calibration marker containing four retro-reflective spheres centered at points $\mathbf{P}_i$, where $i=1:4$, is placed within the C-arm's imaging frustum during X-ray scans. For the  $n$-th scan, the pose of the calibration marker ($Cali$) with respect to Surgeon ($S$), denoted as $T^{Cali}_{S,n} = (R^{Cali}_{S,n}, \mathbf{t}^{Cali}_{S,n})$ ($R \in SO(3)$ is a rotation matrix, and $t \in \mathbb{R}^3$ is the translation vector), and the pose of the C-arm gantry marker ($C$), denoted as $T^{C}_{S,n} = (R^{C}_{S,n}, \mathbf{t}^{C}_{S,n})$, are recorded. Subsequently, the sphere coordinates on the images, $\mathbf{p}_{i,n}$ for $i=1:4$, are located using the Hough transform. The desired transformation at scan $n$, $T_{C,n}^{Source}$, is then calculated by solving the following optimization problem:
\begin{equation}
\min_{R_n, t_n} \sum_{i=1}^{4} \left\| \mathbf{p}_{i,n} - K \left( R_n^{-1} \left(R^{Cali}_{S,n}\mathbf{P}_i + \mathbf{t}^{Cali}_{S,n} \right) - R_n^{-1} \mathbf{t}_{n} \right) \right\|^2
\end{equation}
\begin{equation}
T_{C,n}^{Source} = \left(\left(R^{C}_{S,n}\right)^{-1} R_{n}, \left(R^{C}_{S,n}\right)^{-1}\left(\mathbf{t}_{n}- \mathbf{t}^{C}_{S,n}\right)\right)
\end{equation}
The transformation $T_{C}^{Source}$ is then determined by averaging over $N$ scans.

\subsubsection{Surgical Tool and Insertion Pathway}\label{section:tool}
\noindent The outcome of screw placement critically depends on the insertion pathways created during the procedure. These pathways are typically established using probes or cannulas, into which K-wires are drilled to ensure precise screw placement. To enhance this process, \textit{X-GuideAR} employs a novel cannula design that improves pelvic screw placement by mitigating wire deflection and bending, especially in percutaneous procedures \cite{zhang2024straighttrack}. This design features a rigid sheath that maintains the K-wire's trajectory, even against tissue resistance or irregular bone surfaces, thereby minimizing misalignment and reducing both surgical time and radiation exposure.

To ensure optimal performance, the tool undergoes a one-time pivot calibration utilizing a Polaris system (Northern Digital Inc., Waterloo, ON, Canada) and reflective markers to accurately align and validate the K-wire’s trajectory. This pivot calibration yields $\mathbf{P}_{tip}$, the 3D coordinate of the tool tip in the tool frame, and $\mathbf{n}_{tool}$, a normal vector representing the cannula’s pathway. This calibration process ensures precise intra-operative guidance for the surgeon.

\subsubsection{Pre- and Intra-operative Image Alignment}
\noindent Prior to surgery, CT scans are typically performed to assess the patient's anatomy and facilitate pre-operative planning. Achieving accurate alignment between the pre-operative CT data ($Pre$) and the patient anatomy under intra-operative tracking system ($P$) is crucial for generating reliable DeepDRRs. If the tracking marker is not attached to anatomy during CT scan, this alignment is usually established through 2D/3D registration between intra-operative X-ray images and the CT scans. Grupp et al. proposed a robust registration approach that uses one or more intra-operative X-ray images to estimate the transformation aligning the patient’s current anatomy with the pre-operative CT volume \cite{grupp2020pose}.

\subsection{DeepDRR Generation Module}\label{section:drr generation}
\noindent The DeepDRR framework leverages deep learning models to enhance both the speed and quality of DRR generation, resulting in more realistic and detailed images suitable for efficient real-time use during surgical procedures \cite{mathias2018deepdrr}. In \textit{X-GuideAR}, the DeepDRR images are projected using a projection matrix defined as:

\begin{equation}
P = K \begin{bmatrix} R_{u}^{-1} \mid -R_{u}^{-1}\mathbf{t}_{u} \end{bmatrix}
\label{projection:drr}
\end{equation}
where,
\begin{equation}
T_{u} = (R_u,\mathbf{t}_u) = \left( T_P^{Pre} \right)^{-1} \left( T_S^P \right)^{-1} T_S^C T_C^{\text{Source}}
\label{kinematic:drr}
\end{equation}

\subsection{X-ray Augmentation Module}\label{sec:X-ray Augmentation Module}
\noindent The spatial awareness module allows for the overlay of surgical tools onto X-rays, providing a real-time preview of tool alignment without radiation exposure. Assuming an X-ray is taken with the C-arm pose $T_{S,0}^{C}$, a point $\mathbf{P_0}$ in the tool frame at time $\theta$ can be projected onto the X-ray image as a 2D point $\mathbf{p}_0$ using:
\begin{equation}
\mathbf{p}_0 = K \begin{bmatrix} R_{\theta}^{-1} \mid -R_{\theta}^{-1}\mathbf{t}_{\theta} \end{bmatrix} \mathbf{P_0}
\label{projection:tool}
\end{equation}
where,
\begin{equation}
T_{\theta} = (R_\theta,\mathbf{t}_\theta) = (T_S^T )^{-1} T_S^C T_C^{Source}
\label{kinematic:tool}
\end{equation}

And by having the surgical tool pose, the tool can be projected onto the X-rays: 
\begin{equation}
\begin{split}
    \mathbf{p}(s) = K \begin{bmatrix} R_{\theta}^{-1} \mid -R_{\theta}^{-1}\mathbf{t}_{\theta} \end{bmatrix}
    \left( \mathbf{P}_{tip} + s \mathbf{n}_{tool} \right)
\end{split}
\label{projection}
\end{equation} 
where $\mathbf{P}_{tip} + s \mathbf{n}_{tool}$ represents a point on the axis of cannula.

\subsection{Visualization and Interaction for \textit{X-GuideAR} S2AI Procedure}\label{section:xguidearS2AI}
\noindent \textit{X-GuideAR} aids the S2AI procedure by providing real-time previews for fluoro-hunting and screw navigation. The fluoro-hunting preview reduces radiation exposure during AP, inlet, and tear-drop view acquisition (see Figure \ref{fig:systemoverview}), while the screw navigation preview further minimizes radiation by eliminating X-rays during drill alignment. Following the steps in Figure \ref{fig:systemoverview} and Section \ref{spatialawareness}:
\begin{enumerate}
    \item Surgeon performs fluoro-hunting by moving the C-arm and judging the DeepDRR preview. Once the preview is satisfactory, the C-arm is fixed. 
    \item An X-ray scan is taken with the C-arm stationary, and if accepted, it is combined with $T_{\theta}$ to form a frustum.
    \item Step one and two are repeated for AP, inlet, and tear-drop views, creating three frustums.
    \item The surgeon adjusts the entry point and pivots the cannula while viewing the virtual trajectories projected onto the acquired views. The surgeon proceeds with advance the drill once alignment is satisfactory.
\end{enumerate}
Post drilling X-rays may be followed to check the quality of pathway. Once confirmed, surgeon inserts the S2AI screws along the wire into the pathway, and advance the screw with desired screw length.

\begin{figure*}[t]
    \centering
    \begin{minipage}[b]{0.58\textwidth}
        \centering
        \includegraphics[width=\textwidth]{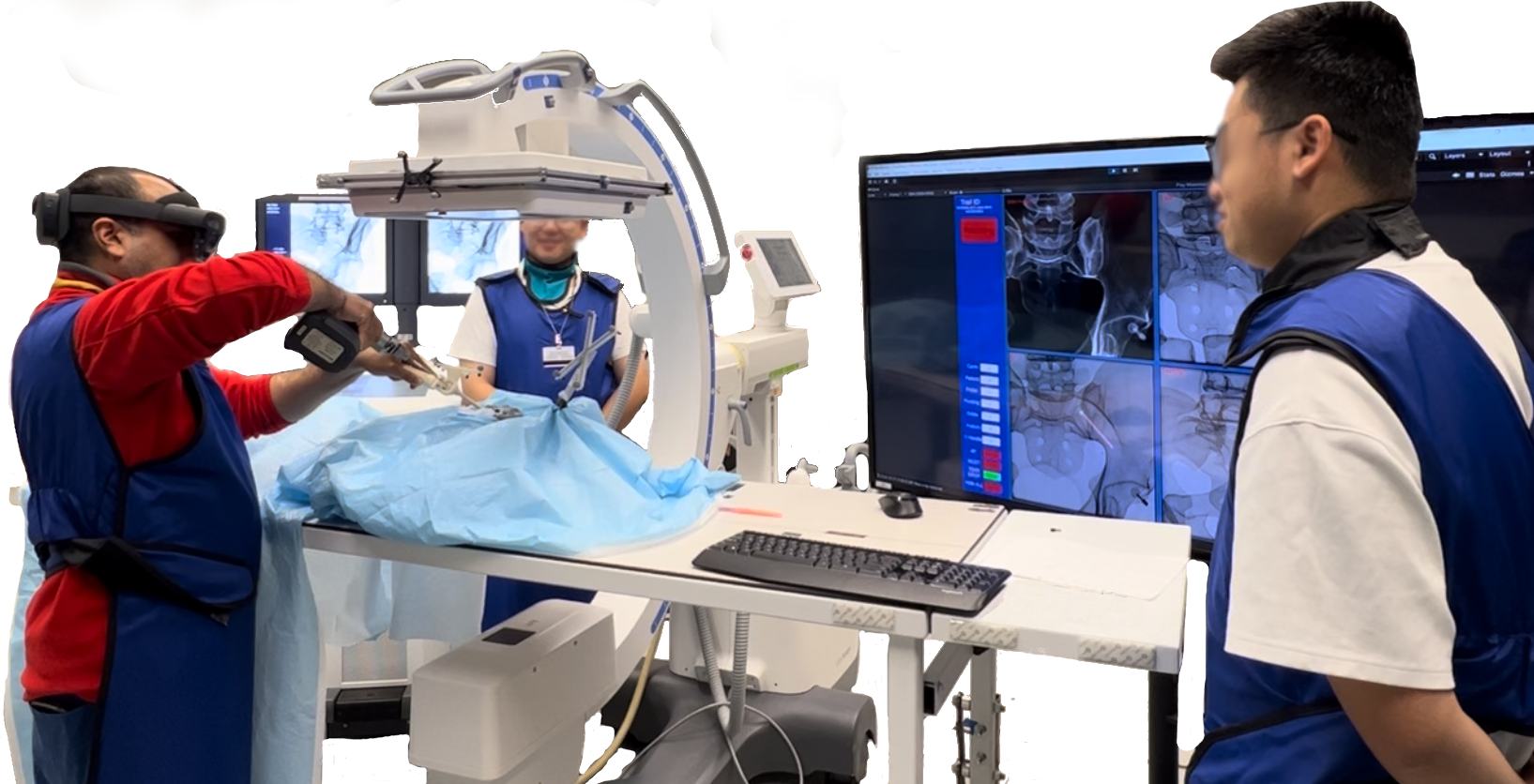}
        % \caption*{}
        \label{fig:mockorexperiment1}
    \end{minipage}
    \hfill
    \begin{minipage}[b]{0.4\textwidth}
        \centering
        \includegraphics[width=\textwidth]{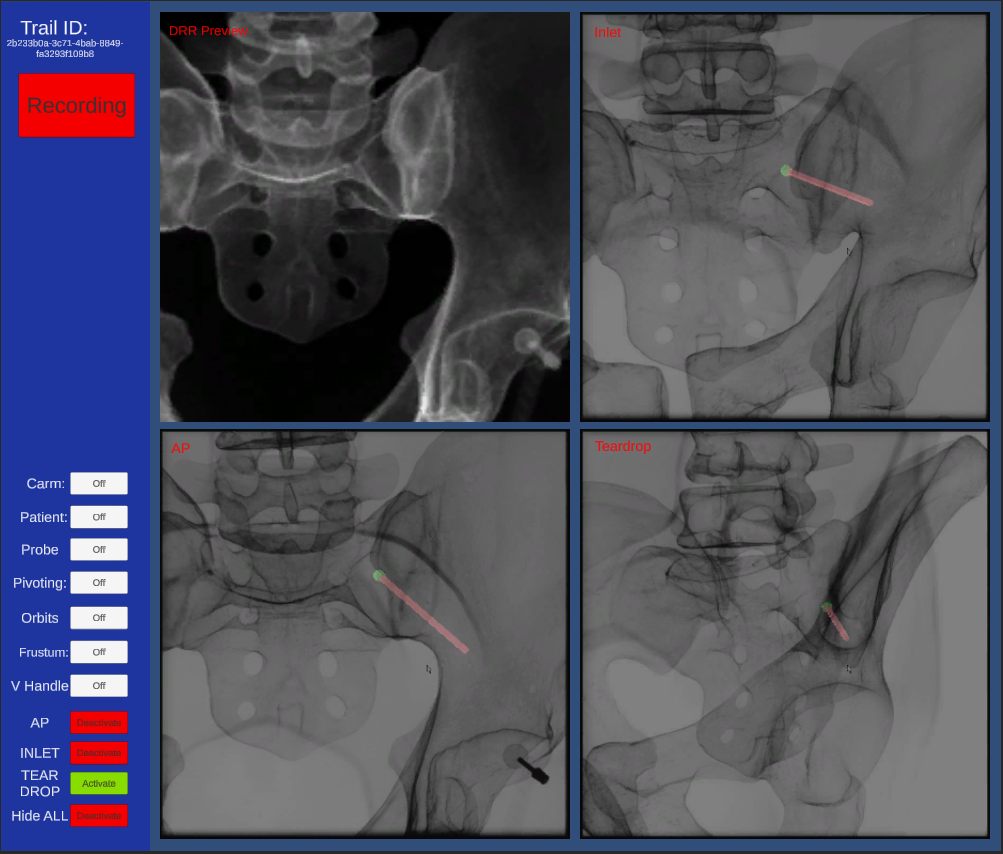}
        % \caption*{}
        \label{fig:mockorexperiment2}
    \end{minipage}
    \caption{
(Left) Experimental setup for the S2AI phantom study, featuring the Siemens Cios Fusion, two displays, and two assistants (white shirts)—one handling C-arm positioning and the other note-taking. The surgeon (red shirt), wearing a HoloLens 2, performs drilling guided by \textit{X-GuideAR}, holding the cannula and drill. (Right) User interface screenshots show buttons on the left for enabling/disabling 3D guidance and selecting fluoro-hunting views, with the four panels on the right displaying DeepDRR previews and X-rays augmented with the tool trajectory.}
    \label{fig:mockorexperiment}
\end{figure*}

\section{Experiment and Result}
\noindent To showcase the value/potential of X-GuideAR, we conducted a preliminary phantom study involving an expert orthopedic surgeon. The experiment included four trials across two methods: traditional S2AI as control groups (trial one and four), and S2AI screw insertion using \textit{X-GuideAR} (trial two and three).

\subsection{Phantom Preparation}
\noindent Each trial utilized a distinct, radio-opaque male pelvic model from Sawbones. For trial two and three, CT scans were acquired. Retro-reflective markers were rigidly affixed to the phantom prior to the acquisition of the CT. This setup served to isolate sources of error due to 2D/3D registration and the proposed workflow. The pelvis was secured in a prone position on a Sawbones holder and covered with a surgical drape, with an incision made above the sacrum region to simulate a clinical scenario.
\subsection{Hardware and Tools}
\noindent The experiment utilized a Siemens Healthcare Cios Fusion as the X-ray device and a Microsoft HoloLens 2 for inside-out tracking and 3D guidance (see Figure \ref{fig:mockorexperiment}). The surgical setup included a 3D-printed cannula (detailed in Section \ref{section:tool}), 2.8 mm × 300 mm orthopedic drill bits, and an orthopedic drill for pathway creation. A server equipped with dual NVIDIA RTX 4090 GPUs was employed for DeepDRR generation, while a Magewell frame grabber streamed real-time X-rays from the Cios Fusion. All DeepDRRs and X-rays were sent to a workstation that provided the graphical user interface for visualization and interaction. A one-time calibration of the Cios Fusion, following the method described in Section \ref{section:X-raysource}, was conducted using an NDI Polaris system. We note that the NDI tracker was used solely for this calibration step and was not used during the subsequent phantom studies. 9 X-rays were captured for the calibration. The registration process achieved a reprojection error of 1.74 mm. 

\subsection{Phantom Studies}\label{section:phantomstudy}
\noindent For each trial, the surgeon aimed to create insertion pathways for S2AI screws on both the right and left sides of the pelvis in a mock operating room setting. As depicted in Figure \ref{fig:mockorexperiment}, the surgeon stood at the operating table, facing the phantom with the patient’s head positioned to the surgeon’s left. The C-arm was initially placed within the surgeon's reach for adjustments. Two assistants participated in the procedure: one stationed behind the C-arm to reposition it as directed and the other at the table to record data. Two digital displays were employed: the left monitor served as the C-arm display, presenting X-ray images, while the right-side monitor, connected to the server, displayed DeepDRRs and augmented X-ray scans.

\subsubsection{Control Phantom Study}\label{S2AI Phantom Study} 
The procedure involved the surgeon directing an assistant to position the C-arm for acquiring X-rays at each required view and guiding the drill to create the insertion pathways. After drilling, X-ray images were taken to verify that the final pathways matched the surgeon's expected trajectories.

The time required to create each pathway and the total number of X-ray shots taken for both sides were recorded as summarized in Table \ref{tab:xray_shots_guideAR}. Table \ref{tab:xray_shots_guideAR} reports the net values, excluding anomalies such as system restarts from software freezes or low-quality X-rays caused by metal interference from the surgical table frame—conditions not expected in a standard operating room workflow.

\subsubsection{X-GuideAR Phantom Study} 
In the X-GuideAR phantom study, the surgeon followed the protocols outlined in Section \ref{section:xguidearS2AI}. The procedure began with the surgeon observing real-time DeepDRRs displayed on the monitor and directing an assistant to adjust the C-arm until the AP view was achieved. The surgeon then captured an X-ray and assessed whether the view met expectations; if not, the aforementioned steps were repeated. This process was conducted for the AP, inlet, and teardrop views.

Once satisfactory X-rays from all three views were obtained, the surgeon used the augmented X-rays to guide the pathway creation, drilling through the cannula sheath after confirming the trajectory. Finally, the assistants repositioned the C-arm as directed to verify the drilling result. The operation time and number of X-ray shots, as summarized in Table \ref{tab:xray_shots_guideAR} for the \textit{X-GuideAR} phantom study, were recorded using the same criteria described in Section \ref{S2AI Phantom Study}.

\subsection{Post-operative Assessments}
\noindent Post-operative CT scans were performed for all four phantoms using the Brainlab Loop-X system. To visualize the pathways, 2 mm pencil leads were inserted into the pathways, creating high-intensity signals in the radiographic images. The CT scans were then imported into 3D Slicer \cite{pieper20043d} for segmentation and annotation. A 2 mm-thick shell model of the outer pelvic surface was generated to mimic the cortical wall. The drill pathways, highlighted by the pencil leads, were annotated.

\begin{figure*}[t]
\centering
\includegraphics[width=0.9\textwidth]{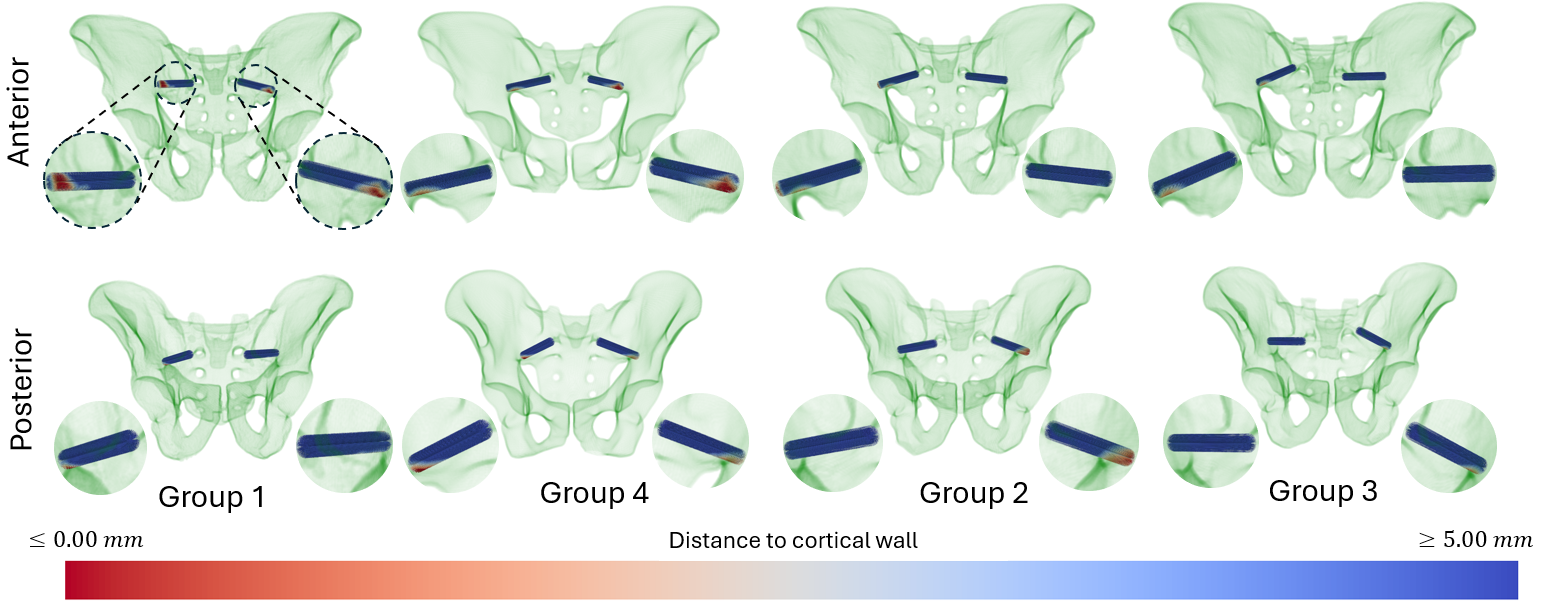}
\caption{Distance-to-cortex maps for virtual $9 mm \times 60 mm$ S2AI screws across the four evaluated trajectories. Anterior and posterior views are shown; color indicates cortical proximity (red $=$ breach, blue $\geq5 mm$ clearance)}\label{fig:tolerancemap}
\end{figure*}

To evaluate the results, two S2AI screws (60 mm and 90 mm in length) were simulated along the annotated pathways, representing the minimum and maximum lengths for achieving reliable biomechanical stability \cite{shin2018effect}. For each trajectory, the closest distance to the ``cortical wall'' determined the maximum allowable screw radius without breaching the wall, which is doubled to be the maximum allowable screw diameters. The first 2 cm of each trajectory, near the sacral surface, were excluded from calculations to prevent skewing the results. The outcomes are summarized in Table \ref{tab:xray_shots_guideAR}. For visualization, virtual 9mm$\times$60mm screws were modeled as cylinders and inserted into the pelvic anatomy, as illustrated in Figure \ref{fig:tolerancemap}.

\begin{table*}[t]
\caption{Experimental results from phantom studies. The table reports the number of fluoroscopic images acquired per anatomical view, procedure time, and the maximum screw diameter that can be placed within the bony corridor without cortical breach.}\label{tab:xray_shots_guideAR}
\centering
\resizebox{\textwidth}{!}{%
\begin{tabular}{|ccc|cccc|c|cc|}
\hline
\multicolumn{1}{|c|}{\multirow{2}{*}{\textbf{Method}}}    & \multicolumn{1}{c|}{\multirow{2}{*}{\textbf{Trial}}} & \multirow{2}{*}{\textbf{Side}} & \multicolumn{4}{c|}{\textbf{Number of X-rays Scanned}}                                                                            & \multirow{2}{*}{\textbf{Total Time}} & \multicolumn{2}{c|}{\textbf{Maximum Screw Diameter with}}                     \\ \cline{4-7} \cline{9-10} 
\multicolumn{1}{|c|}{}                                    & \multicolumn{1}{c|}{}                                &                                & \multicolumn{1}{c|}{\textbf{AP}} & \multicolumn{1}{c|}{\textbf{Inlet}} & \multicolumn{1}{c|}{\textbf{Tear drop}} & \textbf{Total} &                                      & \multicolumn{1}{c|}{\textbf{a 60-mm-Long Screw}} & \textbf{a 90-mm-Long Screw} \\ \hline
\multicolumn{1}{|c|}{\multirow{4}{*}{Regular}}            & \multicolumn{1}{c|}{\multirow{2}{*}{Trial 1}}        & Right                          & \multicolumn{1}{c|}{5}           & \multicolumn{1}{c|}{13}             & \multicolumn{1}{c|}{6}                  & 24             & 256s                                 & \multicolumn{1}{c|}{6.60 mm}              & 6.60 mm              \\ \cline{3-10} 
\multicolumn{1}{|c|}{}                                    & \multicolumn{1}{c|}{}                                & Left                           & \multicolumn{1}{c|}{1}           & \multicolumn{1}{c|}{9}              & \multicolumn{1}{c|}{7}                  & 17             & 360s                                 & \multicolumn{1}{c|}{4.45 mm}              & 4.45 mm              \\ \cline{2-10} 
\multicolumn{1}{|c|}{}                                    & \multicolumn{1}{c|}{\multirow{2}{*}{Trial 4}}        & Right                          & \multicolumn{1}{c|}{6}           & \multicolumn{1}{c|}{5}              & \multicolumn{1}{c|}{6}                  & 17             & 187s                                 & \multicolumn{1}{c|}{2.96 mm}              & 1.84 mm              \\ \cline{3-10} 
\multicolumn{1}{|c|}{}                                    & \multicolumn{1}{c|}{}                                & Left                           & \multicolumn{1}{c|}{9}           & \multicolumn{1}{c|}{5}              & \multicolumn{1}{c|}{5}                  & 19             & 215s                                 & \multicolumn{1}{c|}{9.59 mm}              & 0.47 mm              \\ \hline
\multicolumn{3}{|c|}{Average}                                                                                                                     & \multicolumn{1}{c|}{5.25}        & \multicolumn{1}{c|}{8}              & \multicolumn{1}{c|}{6}                  & 19.25          & 254.5s                               & \multicolumn{1}{c|}{5.9 mm}               & 3.34  mm             \\ \hline
\multicolumn{1}{|c|}{\multirow{4}{*}{\textit{X-GuideAR}}} & \multicolumn{1}{c|}{\multirow{2}{*}{Trial 2}}        & Right                          & \multicolumn{1}{c|}{2}           & \multicolumn{1}{c|}{2}              & \multicolumn{1}{c|}{4}                  & 8              & 294s                                 & \multicolumn{1}{c|}{16.35 mm}             & 15.57 mm             \\ \cline{3-10} 
\multicolumn{1}{|c|}{}                                    & \multicolumn{1}{c|}{}                                & Left                           & \multicolumn{1}{c|}{3}           & \multicolumn{1}{c|}{3}              & \multicolumn{1}{c|}{3}                  & 9              & 343s                                 & \multicolumn{1}{c|}{6.73 mm}              & 2.13 mm              \\ \cline{2-10} 
\multicolumn{1}{|c|}{}                                    & \multicolumn{1}{c|}{\multirow{2}{*}{Trial 3}}        & Right                          & \multicolumn{1}{c|}{2}           & \multicolumn{1}{c|}{2}              & \multicolumn{1}{c|}{2}                  & 6              & 149s                                 & \multicolumn{1}{c|}{18.59 mm}             & 3.96 mm              \\ \cline{3-10} 
\multicolumn{1}{|c|}{}                                    & \multicolumn{1}{c|}{}                                & Left                           & \multicolumn{1}{c|}{2}           & \multicolumn{1}{c|}{2}              & \multicolumn{1}{c|}{2}                  & 6              & 283s                                 & \multicolumn{1}{c|}{10.13 mm}             & 10.13 mm             \\ \hline
\multicolumn{3}{|c|}{Average}                                                                                                                     & \multicolumn{1}{c|}{2.25}        & \multicolumn{1}{c|}{2.25}           & \multicolumn{1}{c|}{2.75}               & 7.25           & 267.25s                              & \multicolumn{1}{c|}{12.95 mm}             & 7.95 mm              \\ \hline
\end{tabular}%
}
\end{table*}

\section{Discussion and Conclusion}
\noindent Based on our preliminary findings, \textit{X-GuideAR} reduced the number of X-rays required for the S2AI procedure by 62.3\% compared with the conventional fluoroscopic guidance workflow (regular group in Table \ref{tab:xray_shots_guideAR}), highlighting its effectiveness in minimizing radiation exposure for patients and the surgical team. Across the two trials (four trajectories) of \textit{X-GuideAR} drill, a reduction in X-ray usage was observed between Trial 2 and Trial 3, possibly showing increased familiarity with the system. In the second \textit{X-GuideAR} phantom study, the surgeon required only two X-rays per view: one for initial guidance during fluoro-hunting and one for post-placement validation. In three out of the four evaluated trajectories (Trial 2 right; Trial 3 left and right), the workflow was further refined such that just three X-rays in total—one per required view—were used until the pathway was established, with additional imaging reserved only for drill-trajectory quality assurance. 

The average fluoroscopy time per trajectory was 4.5 seconds with \textit{X‑GuideAR}. Although additional imaging may occasionally be required for screw adjustment during surgery, the current results still represent a substantial reduction compared with existing minimally invasive screw placement workflows, in which the average fluoroscopy time per screw exceeds 7.2 seconds \cite{pennington2019evaluation}. We acknowledge that, in addition to intra-operative fluoroscopy, pre‑operative CT imaging also contributes to the patient’s overall radiation exposure. However, while some experienced surgeons are able to perform S2AI screw placement without a CT scan, pre‑operative CT remains widely preferred \cite{park2015free, pennington2019evaluation, sherif2023pelvic}. \textit{X-GuideAR} leverages this existing imaging workflow while substantially reducing intra-operative radiation exposure, providing improved guidance within established clinical practice.

In addition to reducing radiation exposure, \textit{X‑GuideAR} did not significantly increase the average surgical duration compared to the regular workflow, which is encouraging given the initial learning experience associated with tools not traditionally integrated into the surgical workflow, including AR HMDs, inside-out tracking, and the cannula. On average, the durations for each operational phase were as follows: fluoro-hunting took 123 seconds, cannula manipulation 84 seconds, and drilling 68.25 seconds. Notably, the time spent on cannula manipulation and drilling was longer than expected. This is partly due to the small tracking marker on the cannula, which results in a narrower tracking field and therefore demands additional attention during initial use. In addition, the cannula requires the surgeon to operate the drill with one hand while stabilizing the cannula with the other, contributing to the learning curve. These factors, which initially prolong the surgical workflow, are expected to diminish as the surgeon becomes more familiar with the system—particularly with cannula control. Of note, pathway creation on the left side consistently took longer than on the right, likely due to the more challenging anatomical access in the experimental setup (i.e., the surgeon was always positioned on the left-hand side of the patient). However, this variability may also depend on individual surgical techniques and preferences.

\textit{X-GuideAR} demonstrated performance consistent with state-of-the-art systems that assist with radiation reduction during X-ray acquisition. Killeen et al. reported an average reduction of approximately 66\% in X-ray usage using a mixed-reality interface for robotic C-arm positioning \cite{killeen2023mixed}. Similarly, Silva et al. showed a reduction of approximately 63\% in the number of fluoroscopic exposures \cite{de2018virtual}. The results of \textit{X-GuideAR}, with an averaged reduction of $62.3\%$, demonstrate consistent efficiency in minimizing X-ray usage, while extending the reduction to the entire procedure and avoiding the need for robotic X-ray systems.

Beyond reducing radiation exposure, \textit{X-GuideAR} demonstrated promising screw trajectory quality. The average maximum safe diameters were 12.95 mm for 60-mm screws and 7.95 mm for 90-mm screws, indicating that standard clinical S2AI implant sizes (7.5–9.5 mm) could be accommodated with adequate bony containment. This allows for the placement of larger diameter screws comparing to the regular guidance, contributing to improved fixation strength and stability, which may lead to better overall surgical outcomes \cite{shin2018effect, wu2017technique}. As shown in Figure \ref{fig:tolerancemap}, the screw pathways maintained safer distances from critical anatomical structures, such as the sciatic notch, while extending more medially into the iliac wall. This increased tolerance reduces the risk of nerve and vessel injuries \cite{arora2022challenges}.

In comparison, Shillingford et al. reported that a free-hand S2AI technique had an 8.5\% cortical breach rate with a mean breach distance of 3.9 mm \cite{ShillingfordJamalN2018HvRA}. The same study showed robotic guidance achieved a lower 4.3\% breach rate. More recent robot-guidance studies reported breach-free outcomes in their analysis cohorts \cite{ good2022robotic}. Intra-operative CT–based navigation has also demonstrated reliable accuracy with implants averaging 81.4 mm × 7.9 mm, larger than those placed under fluoroscopic guidance (72.7 mm × 7.4 mm) \cite{sullivan2023sacropelvic}. Our preliminary study suggests that \textit{X-GuideAR} could achieve comparable trajectory safety to these established approaches. 

The sources of error that may affect the quality of screw placement can be identified twofold: systematic error and human-related error when performing drilling. Specifically, for systematic error, we adapted a Perspective-n-Points algorithm from \cite{fotouhi2019interactive}, resulting in a reprojection error of 1.74 mm. The pivot calibration was tested to have a 0.48 mm RMS error \cite{zhang2024straighttrack}. The HMD tracking has sub-millimeter tracking accuracy \cite{martin2023sttar}. Further post-operative evaluation on a larger sample size with end-to-end analysis could contribute to further validating the significance of this study.

The above-mentioned improvements are expected to be even more significant for less-experienced clinicians or trainees. Compared to existing AR-based methods for the S2AI procedure \cite{heining2024augmented, judy2023human}, \textit{X-GuideAR} provides dynamic and immediate feedback through real-time anatomical representation from the fluoroscopic imager. This is consistent with current clinical practice for most surgeons and medical centers, where fluoroscopy is routinely used to verify screw placement, ensure surgical safety and accuracy, mitigate risks arising from registration or tracking errors, and enhance surgeon confidence during the procedure. Furthermore, although the system incorporates a HoloLens 2 headset, optical tracking, and fiducial markers, it avoids reliance on large robotic platforms or intra-operative CT systems, which may improve practicality and adoption in operating environments.

The spatial awareness feature of \textit{X‑GuideAR} further enhances usability through the integration of the HMD, which improves tracking robustness by maintaining alignment between the tracking sensors and the surgeon’s line of sight, thereby increasing the stability of the DeepDRR preview and tool‑trajectory visualization. In addition, the HMD displays the X-ray field-of-view and tool-trajectory preview in 3D, providing extra guidance for understanding the current fluoroscopic field of view relative to the patient and how the virtual drilling trajectory is aligned. Moreover, integrating the HMD into the workflow could further reduce X-ray usage by enabling a \textit{review} function that allows surgeons to revisit previously established poses without additional fluoro-hunting \cite{fotouhi2019interactive}. This feature, however, was not implemented in the current study. \textit{X-GuideAR} bridges the kinematic chain between the pre-operative CT and the intra-operative scenario, potentially enabling the mapping from pre-operative to intra-operative stage. This integration could benefit \textit{X-GuideAR} in the future to bring surgical plans into the operating room.

The findings of this work should be interpreted in light of several limitations. First, the evaluation was conducted by a single expert surgeon and a limited number of trials, which constrains our ability to assess usability and performance variability across different skill levels. With additional trials, a clearer understanding of the significance of \textit{X-GuideAR} in reducing radiation exposure for both patients and surgeons, the learning curve associated with the system, and its effectiveness among users with varying levels of expertise may be obtained. Second, the experiments were performed on a single phantom model and therefore do not fully capture the anatomical diversity and soft-tissue conditions encountered in clinical practice. Future studies involving more anatomically diverse phantoms or cadaveric specimens will be necessary to strengthen generalizability and clinical relevance.

To conclude, in this work, we introduced \textit{X-GuideAR}, an AR-based surgical copilot system that provides real-time, spatially informed guidance to optimize fluoro-hunting and screw insertion. By generating DeepDRR previews and augmenting X-rays with predictive insertion outcomes, \textit{X-GuideAR} significantly improves radiation safety and insertion accuracy in orthopedic surgeries. Our initial study with an experienced surgeon demonstrated a notable reduction in radiation exposure and an increased allowable screw diameter, indicating improved biomechanical stability and surgical precision. These findings suggest the potential for \textit{X-GuideAR} to enhance surgical outcomes and safety.

\section*{Acknowledgment}

The authors extend their gratitude to Prof. Mathias Unberath and his team members, Mr. Han Zhang and Dr. Benjamin D. Killeen, for their contributions to the design and fabrication of the cannula, as well as for their expertise and insights in DeepDRR.
\bibliographystyle{unsrt}
\bibliography{bibliography}

\end{document}